\pdfoutput=1
\def\year{2021}\relax
\documentclass[letterpaper]{article} 
\usepackage{aaai21}  
\usepackage{times}  
\usepackage{helvet} 
\usepackage{courier}  
\usepackage[hyphens]{url}  
\usepackage{graphicx} 
\usepackage{natbib}  
\usepackage{caption} 
\usepackage{bm}
\usepackage{amsfonts}
\usepackage{multirow}
\title{PGNet: Real-time Arbitrarily-Shaped Text Spotting\\ with Point Gathering Network}
\author{
    Pengfei Wang,\textsuperscript{\rm 1}\thanks{Equal contribution. Fei Qi is the corresponding author.}
    Chengquan Zhang,\textsuperscript{\rm 2}\footnotemark[1]
    Fei Qi,\textsuperscript{\rm 1}\footnotemark[1]
    Shanshan Liu,\textsuperscript{\rm 2}
    Xiaoqiang Zhang,\textsuperscript{\rm 2} \\
    Pengyuan Lyu,\textsuperscript{\rm 2} 
    Junyu Han,\textsuperscript{\rm 2}
    Jingtuo Liu,\textsuperscript{\rm 2}
    Errui Ding,\textsuperscript{\rm 2}
    Guangming Shi\textsuperscript{\rm 1}\\
}
\affiliations{
    \textsuperscript{\rm 1}School of Artificial Intelligence, Xidian University, \\
    \textsuperscript{\rm 2}Department of Computer Vision Technology, Baidu Inc.\\
    pfwang@stu.xidian.edu.cn, zhangchengquan@baidu.com, fred.qi@ieee.org, \{liushanshan07, zhangxiaoqiang01\}@baidu.com, lvpyuan@gmail.com,  \{hanjunyu, liujingtuo, dingerrui\}@baidu.com, gmshi@xidian.edu.cn
}

\begin{document}
\maketitle
\begin{abstract}
The reading of arbitrarily-shaped text has received increasing research attention. However, existing text spotters are mostly built on two-stage frameworks or character-based methods, which suffer from either Non-Maximum Suppression (NMS), Region-of-Interest (RoI) operations, or character-level annotations. In this paper, to address the above problems, we propose a novel fully convolutional Point Gathering Network (PGNet) for reading arbitrarily-shaped text in real-time. The PGNet is a single-shot text spotter, where the pixel-level character classification map is learned with proposed PG-CTC loss avoiding the usage of character-level annotations. With a PG-CTC decoder, we gather high-level character classification vectors from two-dimensional space and decode them into text symbols without NMS and RoI operations involved, which guarantees high efficiency. Additionally, reasoning the relations between each character and its neighbors, a graph refinement module (GRM) is proposed to optimize the coarse recognition and improve the end-to-end performance. Experiments prove that the proposed method achieves competitive accuracy, meanwhile significantly improving the running speed. In particular, in Total-Text, it runs at 46.7 FPS, surpassing the previous spotters with a large margin.
\end{abstract}

\section{Introduction}
Recently, scene text reading has attracted extensive attention in both academia and industry for its numerous applications, such as scene understanding, image retrieval, augmented reality translation~\cite{wu2019editing}, and robot navigation. Thanks to the surge of deep neural networks, significant progress has been made in detection and recognition separable solutions~\cite{wu2017self, long2018textsnake, wang2019shape, wang2019single, Zhan2019CVPR,shi2018aster, yu2020accurate, wan20192dctc},  as well as end-to-end text spotting methods. However, existing end-to-end models~\cite{sun2018textnet, liu2018fots, feng2019textdragon} are mostly built on two-stage frameworks or character-based methods~\cite{xing2019convolutional,yao2018mask} with complex pipelines, which are inefficient for real-time applications. In this paper, we try to investigate a real-time text spotter for arbitrarily-shaped text.

\begin{figure}
    \centering
    \includegraphics[width=0.90\linewidth]{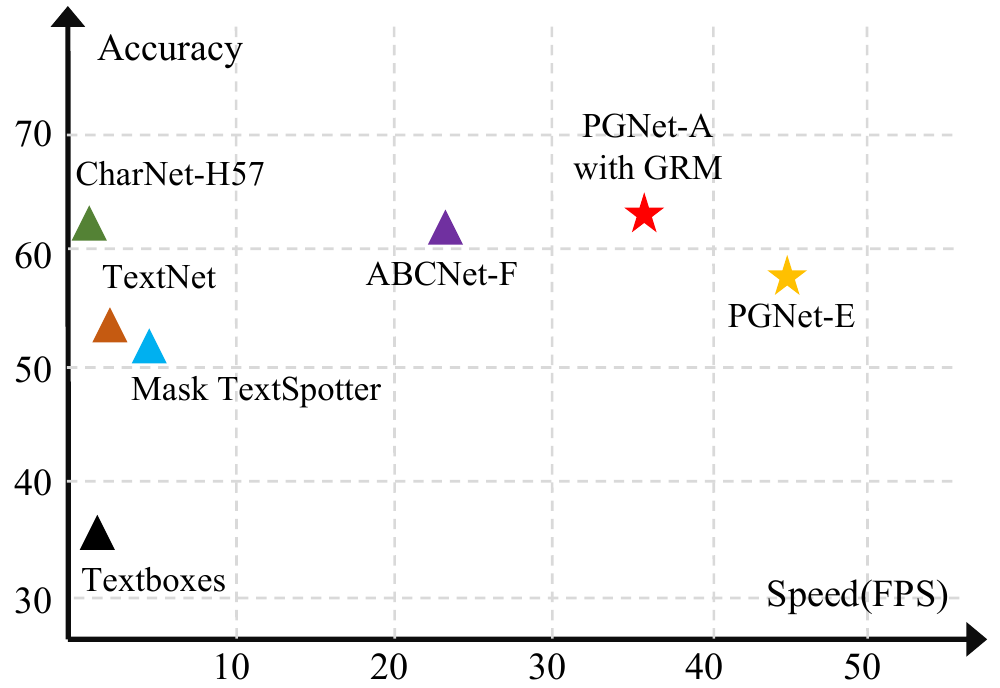}
    \caption{Model Speed vs. Recognition Accuracy on Total-Text: Our PGNet-E achieves at least two times faster than the most recent state-of-the-art method ABCNet~\cite{liu2020abcnet} with competitive recognition accuracy. Complete results are in Table.~\ref{tab:tt}.}
    \label{fig:insight}
\end{figure}

\begin{figure*}
    \centering
    \includegraphics[width=\linewidth]{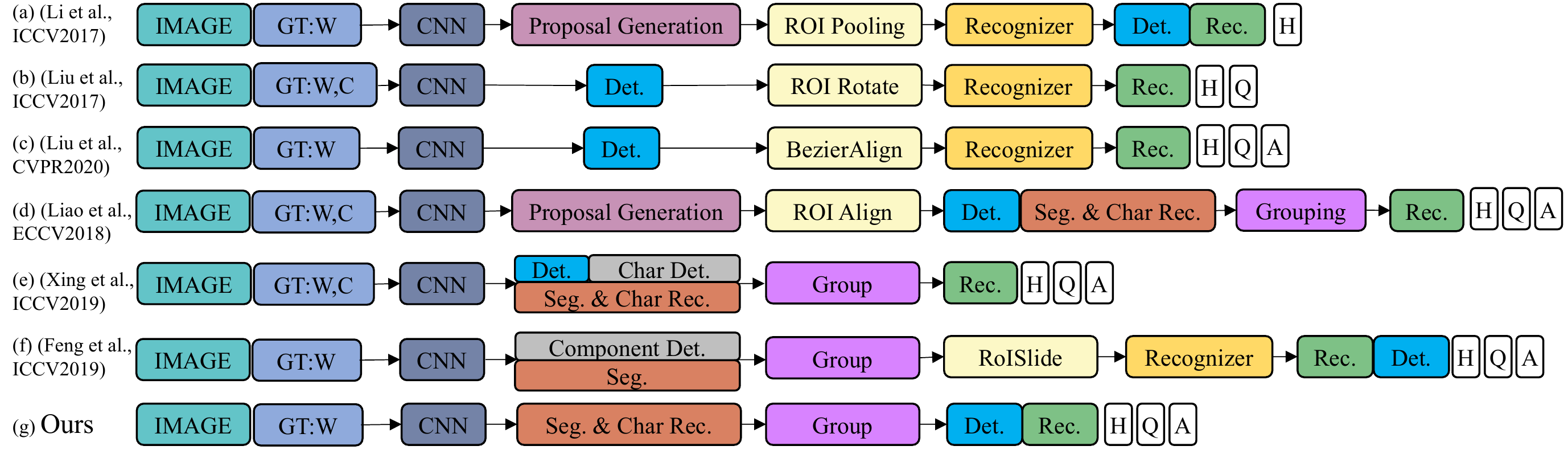}
    \caption{Overview of some end-to-end scene text spotting methods that are most relevant to ours, and the blue and green boxes represent their detection and recognition results. Inside the GT (ground-truth) box, `W' and `C' represent word-level and character-level annotation. The `H', `Q', and `A' represent that the method can detect horizontal, quadrilateral, and arbitrarily-shaped text, respectively. Our method is free from character-level annotations, NMS, and RoI operations.}
    \label{fig:compare}
\end{figure*}

Reading of arbitrarily-shaped scene text is a challenging task, as compared in Fig.~\ref{fig:compare}, and the most recent works may suffer from the following disadvantages: (1) The pipelines of two-stage methods~\cite{sun2018textnet, yao2018mask, feng2019textdragon, liu2020abcnet} are inefficient, which may involve time-consuming  Non-maximum Suppression (NMS) and Region of Interest (RoI) operations. Especially for arbitrarily-shaped text spotter, specific RoI transformation operation, such as RoISlide~\cite{feng2019textdragon} or BezierAlign~\cite{liu2020abcnet}, brings non-negligible computational overhead. (2) In Mask TextSpotter~\cite{yao2018mask} and CharNet~\cite{xing2019convolutional}, character-level annotations are required for training, which is too expensive to afford. Though CharNet could be trained in a weakly supervised manner by character-level annotations in synthetic datasets, free synthesized data is not completely replaceable for real data in practice. (3) The recognition of text in non-traditional reading directions would be failed with pre-defined rules. For example, TextDragon~\cite{feng2019textdragon} and Mask TextSpotter~\cite{yao2018mask} make a strong assumption that the reading direction of text region is either from left to right or from up to down, which precludes correct recognition of more challenging text.

\begin{figure*}
    \centering
    \includegraphics[width=\linewidth]{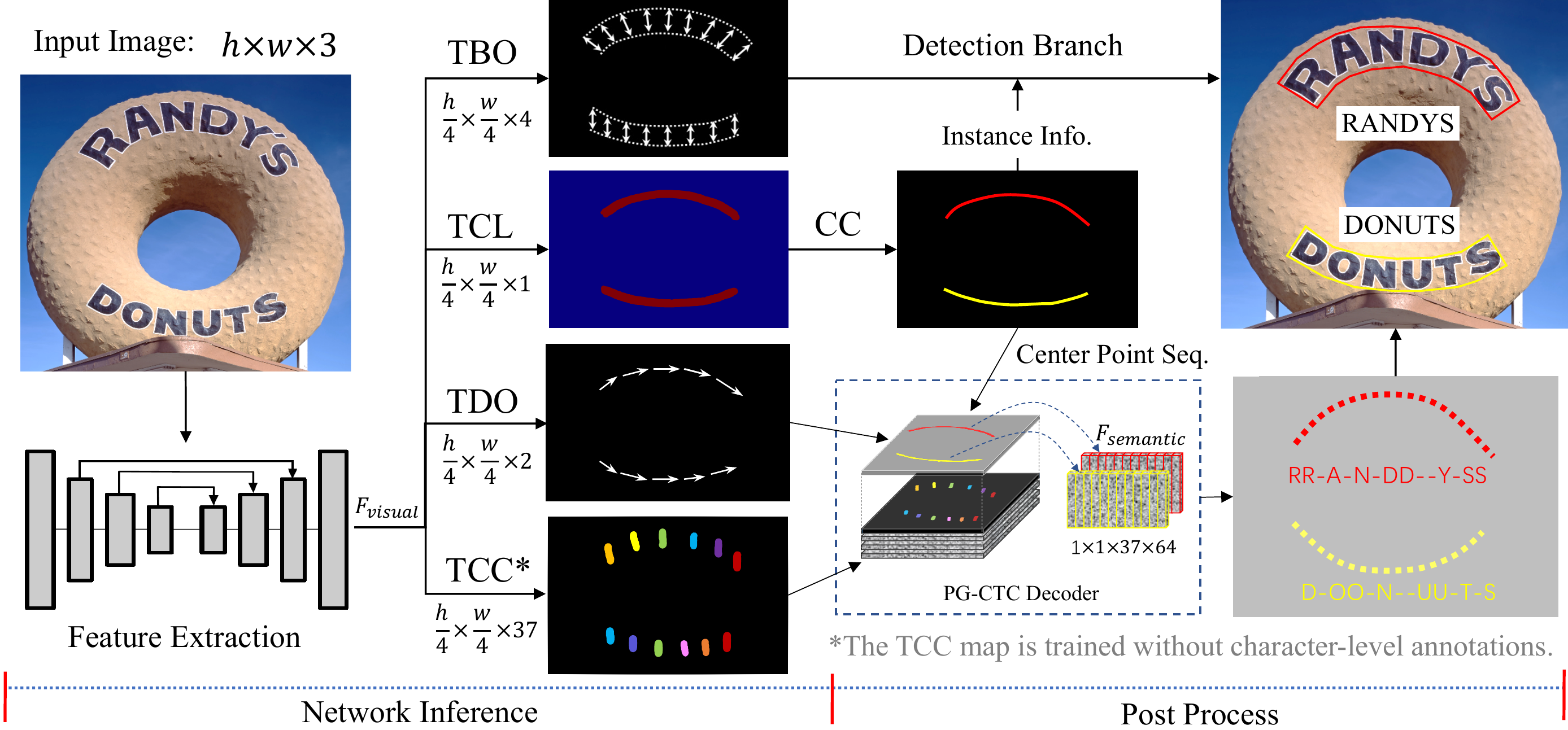}
    \caption{The pipeline of PGNet: 1) Extract feature from an input image, and learn TCL, TBO, TDO, TCC maps as a multi-task problem; 2) The detection and recognition of each text instance can be achieved in a single shot by polygon restoration and PG-CTC decoding mechanism with the center point sequence of each text region.}
    \label{fig:pipeline}
\end{figure*}

In this paper, we propose a novel framework for reading text in real-time speed with point gathering operation, namely PGNet. The PGNet is a single-shot text spotter based on multi-task learning. The architecture of PGNet is shown in Fig.~\ref{fig:pipeline}. We employ an FCN~\cite{milletari2016v} model to learn various information of text regions simultaneously, including text center line (TCL), text border offset (TBO), text direction offset (TDO), and text character classification map (TCC). The pixel-level character classification map is trained with a proposed Point Gathering CTC (PG-CTC) loss, making it free from character-level annotations. In the post-processing, we extract the center point sequence in the reading order of each text instance with TCL and TDO maps, and the detection results can be obtained with the corresponding boundary offset information from TBO map. Using the PG-CTC decoder, we serialize the high-level two-dimensional TCC map to character classification probability vector sequences which can be further decoded to the recognition results. The details will be discussed in Sec.\ref{sect_PG}. As depicted in  Fig.~\ref{fig:compare} and Fig.~\ref{fig:insight}, our pipeline is simple yet efficient, and experiments on public benchmarks prove that PGNet achieves better or competitive performance with excellent running speed.

Moreover, inspired by SRN~\cite{yu2020accurate} and GTC~\cite{hu2020gtc}, we propose a graph refinement module (GRM) to  make secondary reasoning to improve the end-to-end performance further. The points in a text sequence are formulated as nodes in a graph, where the representation of each node is enhanced with semantic context and visual context information from its neighbors, and the character classification result should be more accurate. 

The contributions of this paper are three-fold: 
\begin{itemize}
\item[$\bullet$] We propose a simple yet powerful arbitrarily-shaped text spotter (PGNet), which is free from character-level annotations, NMS, and RoI operations, and it achieves better or competitive performance in end-to-end performance with excellent running speed;
\item[$\bullet$] We introduce a mechanism to restore the reading order of characters in each text instance, making our method able to correctly recognize text in more challenging situations and non-traditional reading directions;
\item[$\bullet$] We also propose an efficient graph refinement module (GRM) to improve the CTC recognition.
\end{itemize}

\section{Related Work}
In this section, we will review some representative scene text spotters, as well as some recent progress in graph neural networks. A comprehensive review of recent scene text spotters can be found in~\cite{ye2015text, zhu2016scene, baek2019wrong, liao2019mask}.

\textbf{Scene Text Spotting.} Inspired by the generic object detection methods~\cite{Liu2016SSDSS,ren2015faster,redmon2016yolo} and segmentation methods~\cite{he2017mask, milletari2016v}, the text spotting methods are developed from spotting regular scene text to spotting arbitrarily-shaped scene text. \citeauthor{lee2016recursive}~\shortcite{lee2016recursive} proposed the first successful end-to-end text recognition model, which only supports horizontal text and requires relatively complex training procedures. To address the multi-orientation problem of text, \citeauthor{busta2017deep}~\shortcite{busta2017deep} utilize YOLO~\cite{redmon2016yolo} to generate rotational proposals, and train RoI sampled features with CTC loss. Inspired by the Faster RCNN~\cite{ren2015faster}, TextNet~\cite{sun2018textnet} generates text proposals in quadrangles, and encodes the aligned RoI features into context information with a simple recurrent neural network to generate the text sequences, which contains some background information; thus, it may suffer from reading curve texts. 

For the spotting of arbitrarily-shaped scene text, Mask TextSpotter~\cite{yao2018mask} detects and recognizes text instances of arbitrary shapes by segmenting the text regions and character regions. However, the character-level annotations are required for training, which is too expensive to afford. Considering the arbitrarily-shaped region of text as a series of quadrangles, TextDragon~\cite{feng2019textdragon} extracts the components of text feature through RoISlide and recognizes each cropped feature with CTC based text recognizer. ABCNet~\cite{liu2020abcnet} attempts to adaptively fit arbitrarily-shaped text by a parameterized Bezier curve to reduce the computation overhead, and they propose a BezierAlign layer for extracting accurate convolution features of a text instance, making it suitable for some real-time scenarios. Mask Textspotter, TextDragon, and ABCNet are capable of spotting arbitrarily-shaped scene text, but they are all RoI-based methods and may involve NMS, RoI cropping, and pooling operations, which is time-consuming and may reduce the performance. 
CharNet is the first one-stage arbitrarily-shaped scene text spotting method, which requires character-level ground truth data for training and its backbone is too heavy to run in a real-time speed.

\textbf{Graph Neural Networks.} The unstructured data is common in many machine learning tasks and can be organized as graphs. Thanks to the previous effort, we can use the convolutional neural network for graph-structured data. The graph convolutional networks (GCNs) can be categorized into spectral methods and spatial methods. Spectral GCNs generalize convolution by Graph Fourier Transform, while spatial GCNs directly perform manually-defined convolution on graph nodes and their neighbors. For more details about GCNs, please refer to~\cite{wu2019comprehensive}. In terms of applications, GCNs is proved to be efficient in many computer vision tasks, including scene graph generation, point clouds classification, and action recognition. For example, \citeauthor{zhang2020det}~\shortcite{zhang2020det} propose a relational reasoning graph network for arbitrary shape text detection by predicting linkages of text components. In this paper, we adopt the Spatial GCN to reasoning the semantic information between point and its neighbors to improve the CTC recognition.

\section{Methodology}
The architecture of our proposed method is shown in Fig.~\ref{fig:pipeline}. Firstly, the input image is fed into a stem backbone with FPN to produce feature $F_{visual}$. Then, the $F_{visual}$ is used to predict TCL, TBO, TDO, and pixel-level TCC map by multi-task learning in parallel at $1/4$ size of the input images. In the training period, the TCL, TBO, and TDO are supervised by the same scale label maps, while a PG-CTC loss is proposed to train the pixel-level TCC map to solve the lack of character-level annotations. In the inference period, we extract the center point sequence of each text instance from TCL, and sort them with TDO information to recover the right reading order, making our method recognize text in non-traditional reading directions correctly.
With the aid of the corresponding boundary offset information from TBO, the detection of each text instance can be achieved in a single shot by polygon restoration. Simultaneously, the PG-CTC decoder can serialize high-level two-dimensional TCC map to character classification probability sequences and decode them into final text recognition results.

\subsection{Point Gathering CTC}
\label{sect_PG}
The point gathering (PG) operation plays an important role in both the training and inference process of PGNet and helps to get rid of character-level annotations, NMS, and RoI operations. The TCC maps of PGNet are maps of 37 characters, including 26 letters, 10 Arabic numerals, and one background class. The point gathering operation is employed to gather the character classification probability sequence from the TCC map according to the center points in the center of each text region, which can be formulated as
\begin{equation}
    P_{\pi} = gather(TCC, \pi),
    \label{eq_gather}
\end{equation}
where $\pi = \{ p_1, p_2, \dots, p_N \}$ is a center point sequence of length $N$, and $p_i = (x_i, y_i)$.  The output $P_{\pi}$ is the character classification probability sequence with size $N \times 37$.

In the training process, the proposed PG-CTC loss makes the training of pixel-level TCC map free from character-level annotations. The typical CTC loss function tackles the training problem of the source and target sequences with inconsistent lengths with a background class. The CRNN~\cite{shi2017crnn} framework transforms the height of the feature map to 1, which may suffer from the background noise while recognizing the curved text. The 2D-CTC~\cite{wan20192dctc} expands the search path of CTC to two-dimensional space, but it still can not handle an image with multi text instances. Here we address the problem by PG-CTC and formulate the classic CTC loss as $CTC\_loss(P, L)$, where $P$ is a character classification probability sequence and $L$ is its transcript label. For an image with ${M}$ text instances, suppose the center point coordinate sequences are $\{\pi_1, \pi_1, ..., \pi_M\}$, and corresponding transcript labels are $\{L_1, L_1, ..., L_M\}$, then we define the PG-CTC loss as
\begin{equation}
L_{PG-CTC} = \sum_{i=1}^M CTC\_loss(P_{\pi_{i}}, L_i),
\end{equation}
where we can calculate the centerline of polygonal word-level annotations and sample it densely to obtain the center point sequences $\pi_i$ in the training process, instead of using character-level annotations. With the training of big data, the character classification information of each pixel in TCC could be learned. 

In the inference process, the PG-CTC decoder dramatically simplifies the overall pipeline of an end-to-end arbitrarily-shaped text spotter, and the NMS and RoI operations are not required in PGNet. For a text region in the TCL map, we extract a center point sequence and sort it in the right reading order, which can be donated as ${\pi}$. Specifically, we adopt a morphological method~\cite{zhang1984fast} to get the skeleton of a text region and treat it as the center point sequence. The text direction of each point can be extracted from TDO maps. We calculate an average direction of all points and sort them according to the length of projection along the direction to obtain the center point sequence ${\pi}$. 
The character classification probability sequence $P_{\pi}$ can be extracted with Eq.~(\ref{eq_gather}), and the PG-CTC decoder can be denoted as 
\begin{equation}
R_{\pi} = CTC\_decoder(P_{\pi}),
\end{equation}
where $R_{\pi}$ represents the transcription of ${\pi}$. For the polygon restoration, we obtain the corresponding border point pairs of ${\pi}$ with TBO maps in the same position, and link all the border points clockwise to obtain a complete polygon representation. For more details about polygon restoration, please refer to our previous SAST~\cite{wang2019single}. Compared with the CTC-based CRNN framework, the PG-CTC could handle images with multi-text instances of arbitrary shape, where the application of CTC loss is expanded a lot. 

\begin{figure}
    \centering
    \includegraphics[width=\linewidth]{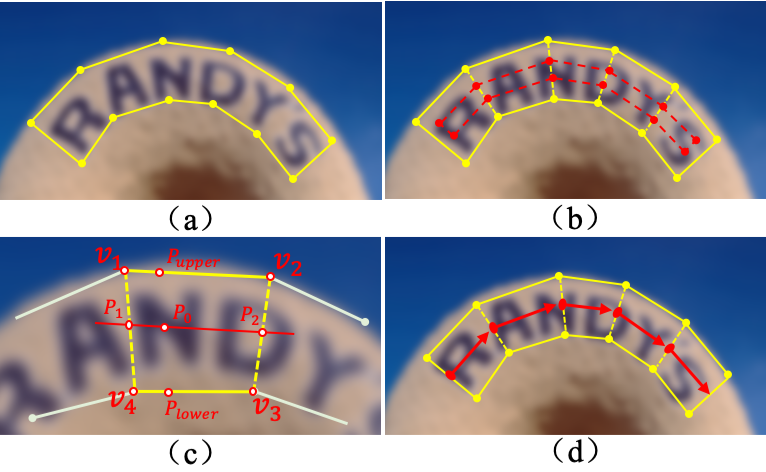}
    \caption{Label Generation: (a) is the ground truth annotation of curved text in yellow; (b)--(d) are the generation of TCL, TBO, and TDO maps, respectively.}
    \label{fig:datagen}
\end{figure}

\subsection{Network Architecture}
Considering the limitation of computing resources in different scenarios, we propose two versions of PGNet, that is, PGNet-Accuracy and PGNet-Efficient, denoted by PGNet-A and PGNet-E in the following sections. The only difference is stem network, and the PGNet-A adopts ResNet-50 as the backbone network, while PGNet-E employs EfficientNet-B0. With different levels of feature map from the stem network gradually merged three-times in the FPN manner, a fused feature map $F_{visual}$ is produced at $1/4$ size of the input images. The TCL and the other maps are predicted in parallel, where we adopt a $1\times 1$ convolution layer with the number of output channels set to \{1, 2, 4, 37 \} for TCL, TDO, TBO, TCC map, respectively. 

\subsection{Label Generation}
The label generation of arbitrarily-shaped text is shown in Fig.~\ref{fig:datagen}. The TCL map is the shrunk version segmentation of the text region. The TBO map indicates the offset between each pixel in TCL and the corresponding point pair in the upper and lower edge of its text region, which helps to determine the boundaries of text regions in the inference. We follow SAST~\cite{wang2019single} to generate TCL and TBO map, where more details are introduced. Inspired by the reading mechanism of humans that the eye moves from one character to the next character along the centerline of text region while reading, the TDO map is estimated to recover the reading order of scene text components, which benefits both detection and recognition tasks, especially for those scene text in non-traditional reading directions. The TDO map indicates the offset vector of each pixel in the TCL map to the next reading position. For a quadrilateral region annotation, the direction of the offset vector is from the center point of the left edge to the center point of the right edge, and its magnitude is the length of text region normalized by the number of characters. Polygonal annotations of more than four vertices are treated as a series of quadrangles connected together, and TBO map and TDO map can be generated gradually from quadrangles as described before.

\subsection{Training Objectives}
The loss of multi-task learning can be formulated as
\begin{equation}
L = {\lambda}_1 L_{tcl} + {\lambda}_2 L_{tbo} + {\lambda}_3 L_{tdo} +{\lambda}_4 L_{tcc},
\end{equation}
where $L_{tcl}$, $L_{tbo}$, $L_{tdo}$ and $L_{tcc}$ represent the loss of TCL, TBO, TDO and TCC maps. We train TCL branch by minimizing the Dice loss~\cite{milletari2016v}, and the Smooth $L_1$ loss~\cite{fastrcnn} is adopted for TBO and TDO map, while TCC map is trained with PG-CTC loss as mentioned before. The loss weights ${\lambda}_1$, ${\lambda}_2$, ${\lambda}_3$, and ${\lambda}_4$ are set to \{1.0, 1.0, 1.0, 5.0\} empirically.

\subsection{Graph Refinement Module}
We also propose a graph refinement module to perceive the word-level semantic context and visual context to improve the end-to-end reading performance with GCNs further. 

\begin{figure}
    \centering
    \includegraphics[width=0.90\linewidth]{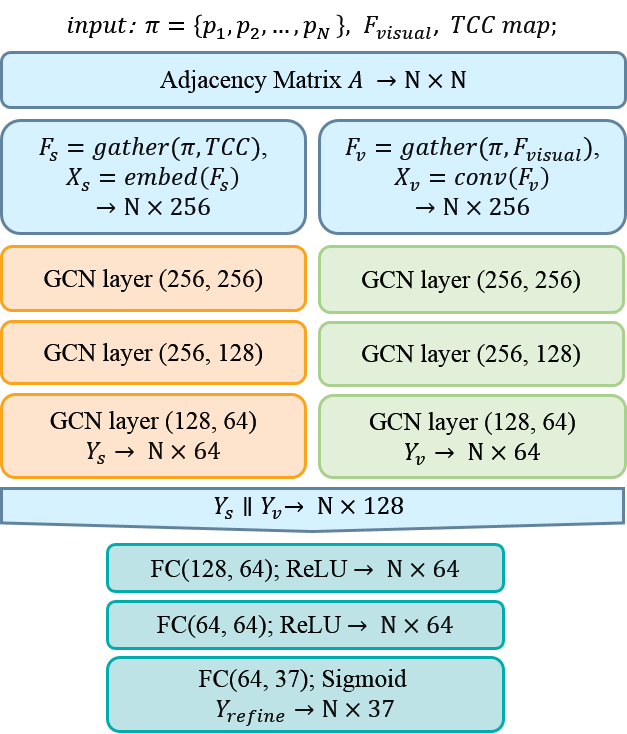}
    \caption{The structure of GRM. For each text sequence, we construct a visual graph and a semantic graph, respectively, and their output ${Y_v}$ and ${Y_s}$ are concatenated together for further character classification with several FC layers.}
    \label{fig:grm_pipeline}
\end{figure}

The coarse recognition results is refined in text instance level, and we construct a visual reasoning graph and a semantic reasoning graph for a point sequence $\pi$, of which the points are considered as the nodes in a graph. We use the same structure of graph convolution layer as~\cite{wang2019gncclust} and~\cite{zhang2020det}. Especially, the $F_{visual}$ and TCC map are both required as inputs. $F_{visual}$ is the output feature of FPN, as illustrated in Fig.\ref{fig:pipeline}. For a point sequence $\pi = \{ p_1, p_2, \dots, p_N \}$, the adjacency matrix is defined as
\begin{equation}
    \bm{A_{ij}}=1 - D(p_i, p_j) / \max(A),
\end{equation}
where $D(p_i, p_j)$ is a L2 distance between $p_i$ and $p_j$, and each node is self-connected. 
The structure of GRM is depicted in Fig.\ref{fig:grm_pipeline}, and the two numbers in brackets indicate $d_{in}$ and $d_{out}$ of a neural layer. In the semantic reasoning graph, the feature map $F_{s}$ is obtained by point gathering operation from TCC map and further embedded to $X_{s}$ of shape $N \times 256$. With three graph layers, the input ${X_{s}}$ is transformed to produce ${Y_{s}}$ of shape $N \times 64$; In the visual reasoning graph, the features $F_{v}$ is gathered from $F_{visual}$, and its channel is transformed from 128 to 256 with several convolutional layers to get $X_{v}$. With a similar network, we can get the visual reasoning output ${Y_{v}}$ of shape $N\times 64$; Finally, we concatenate ${Y_{v}}$ and ${Y_{s}}$, and treat it as a classification problem with several FC layers to produce the refined probability sequence, where the GRM is also optimized with CTC loss. It is worth mentioning that we pad the coarse recognition sequence to the same length and batched together for efficient training, and the max length is set to 64. 

\section{Experiments}

\subsection{Datasets}
The benchmark datasets used for the experiments in this paper are briefly introduced below.

\textbf{ICDAR 2015.} The ICDAR 2015 dataset~\cite{karatzas2015icdar} is collected for the ICDAR 2015 Robust Reading Competition, with 1,000 natural images for training and 500 for testing. The text instances are annotated in word-level.

\textbf{Total-Text.} The Total-Text~\cite{ch2017total} is another curved text benchmark, which consists of 1,255 training images and 300 testing images with multiple orientations: horizontal, multi-Oriented, and curved. 

\subsection{Implementation Details}
\textbf{Training.} The stem network is initialized with pre-trained weight on ImageNet~\cite{deng2009imagenet}. The training process is mainly divided into the warming-up step, fine-tuning step, and training step of the GRM module. In the warming-up step, we apply Adam optimizer to train our model with learning rate 1e-3, and the learning rate decay factor is 0.94 on the SynthText~\cite{gupta2016synthetic}; In the fine-tuning step, the learning rate is re-initiated to 1e-3, and the model is tuned on ICDAR2015 and Total-Text data; The GRM is an additional module in our pipeline, and in the last step, we only train the learnable parameters in GRM module to provide both the metrics on ICDAR 2015 and Total-Text with/without GRM for a fair comparison.

The experiments are performed on a workstation with the following configuration, CPU: Intel(R) Xeon(R) CPU E5-2620; GPU: NVIDIA TITAN Xp $\times 4$; RAM: 64GB. 

\subsection{Ablation Study}
In this section, comprehensive experiments are conducted to prove the strengths of the proposed method.

\textbf{The Effectiveness of TDO.} The TDO map is adopted to recover the sequence information of text in non-traditional reading directions, which benefits both detection and recognition tasks. To verify the efficiency of TDO, we conduct several experiments in ICDAR2015 and Total-Text for PGNet-A and PGNet-E. As shown in Tab.~\ref{tab:ab_tdo}, the benefits of TDO in Total-Text are apparent, and the recognition accuracy in PGNet-E obtains 1.6\% improvement, while it is less than 0.5\% in ICDAR2015. We claim that the non-traditional texts are more likely to appear in the curved Total-Text, while the texts in ICDAR2015 are mostly quadrangles, and the predefined rules may cover most situations.

\textbf{The Effectiveness of GRM.} The GRM is proposed to model word-level semantic context implicitly and improve the end-to-end reading performance. To verify its efficiency, we conduct several experiments in ICDAR2015 and Total-Text on both PGNet-A and PGNet-E, respectively. As depicted in Tab.~\ref{tab:ab_grm}, the efficiency of GRM is proved, and the end-to-end reading performance is improved significantly. Especially, it gains 2.2\% for end-to-end recognition tasks for PGNet-E in Total-Text.

\begin{table}[]
    \centering
    \tabcolsep=4pt\relax
    \begin{tabular}{c|c|c|c|c}
    \hline
    Data Set                            & \multicolumn{2}{c|}{ICDAR2015} & \multicolumn{2}{c}{Total-Text}\\
    \hline
    Method                              & PGNet-A        & PGNet-E       & PGNet-A        & PGNet-E\\
    \hline
    \multicolumn{1}{l|}{with TDO}       & 62.3           & 57.4          & 61.7           & 58.4\\
    \multicolumn{1}{l|}{w/o TDO}        & 61.9           & 56.9          & 60.4           & 56.8\\
    \hline
    Gain                                & +0.4           & +0.5          & +1.3           & +1.6\\
    \hline
    \end{tabular}
    \caption{Evaluation the effectiveness of TDO to improve the recognition performance.}
    \label{tab:ab_tdo}
\end{table} 

\begin{table}[]
    \centering
    \tabcolsep=4pt\relax
    \begin{tabular}{c|c|c|c|c}
    \hline
    Data Set                            & \multicolumn{2}{c|}{ICDAR2015} & \multicolumn{2}{c}{Total-Text} \\ \hline
    Method                              & PGNet-A        & PGNet-E       & PGNet-A        & PGNet-E       \\ \hline
    \multicolumn{1}{l|}{with GRM}       & 63.5           & 58.7          & 63.1           & 60.6          \\
    \multicolumn{1}{l|}{w/o GRM}        & 62.3           & 57.4          & 61.7           & 58.4          \\ \hline
    Gain                                & +1.2           & +1.3          & +1.4           & +2.2          \\ \hline
    \end{tabular}
    \caption{Evaluation the effectiveness of GRM to improve the recognition performance.}
    \label{tab:ab_grm}
\end{table}

\subsection{Evaluation for Curved Text}
On Total-Text, we evaluate the performance of PGNet for spotting scene text of arbitrary shapes. We fine-tune our model with Total-Text training set and partial ICDAR2019 ArT~\cite{icdar19art}, where we remove the images from the Total-Text test set by comparing the MD5 value of each image. The ratio of these two datasets is set to 3: 2 during training. The GRM is further trained with SynthText, partial ICDAR2019 ArT, and Total-Text, of which the data ratio is set to 6: 2: 2, while the learnable parameters outside GRM are frozen. In the inference phase, the longer side of images is resized to 640, with the aspect ratio kept.

The results are shown in Tab.~\ref{tab:tt} and Fig.~\ref{fig:res}. Our method achieves the state-of-the-art detection result and is superior to the other competitors by almost 1.5\%. Besides, with the help of efficient post-processing, our PGNet-A also achieves comparable end-to-end recognition results. Specifically, compared with the previous most accurate method CharNet H-57, our method has almost the same recognition accuracy, but it is over 30 times faster and free from character-level annotations. Compared with the ABCNet, which is the current fastest arbitrarily-shaped text spotter with a speed of 20+ FPS, our PGNet-E runs nearly two times faster than ABCNet-F with better recognition accuracy. The comparisons with the previous state-of-the-arts demonstrate the efficiency and effectiveness of our method.

\subsection{Evaluation for Multi-oriented Text} We also conduct experiments on ICDAR2015 to confirm the superiority of the proposed method on the multi-oriented scene text. Following the same training strategy as Total-Text, The ICDAR2017-MLT-Latin and ICDAR2015 are used in fine-tune stage, of which the data ratio is 3:7, while SynthText, ICDAR2017-MLT-Latin~\cite{nayef2017icdar2017} and ICDAR2015 with a ratio of 6:2:2 are adopted for training GRM. In the test phase, the longer side of images is resized to 1536, with the aspect ratio kept.

\begin{figure*}
    \centering
    \includegraphics[width=\linewidth]{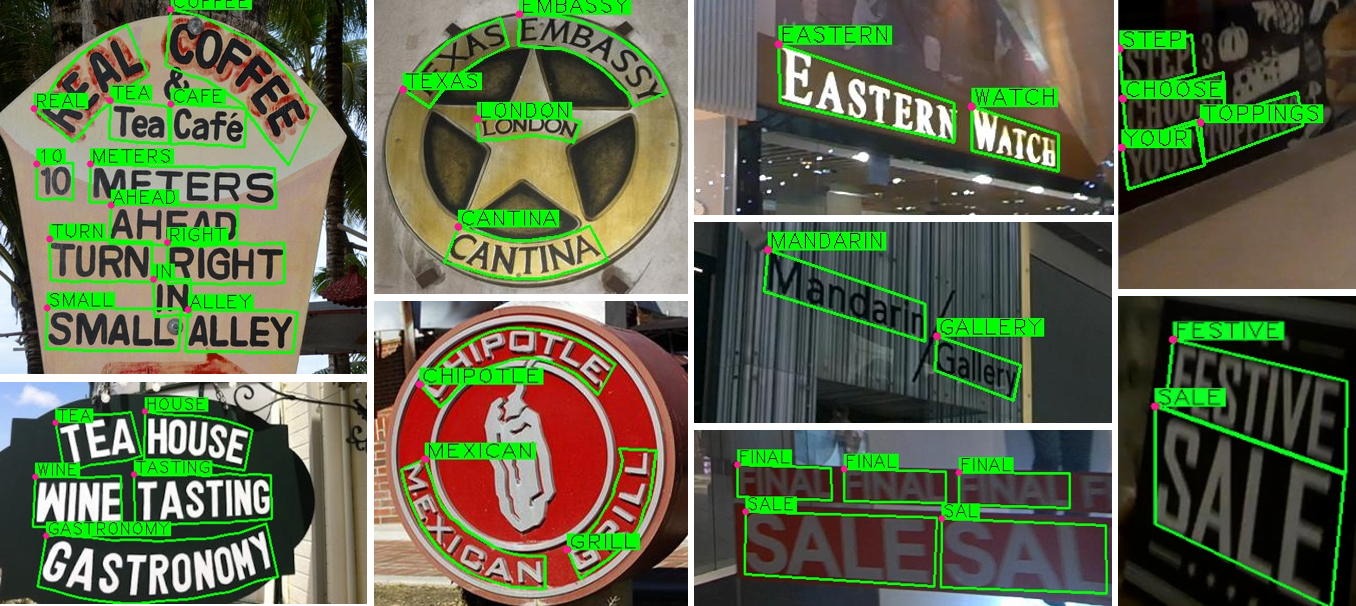}
    \caption{Qualitative results of our method on  Total-Text (left two columns) and ICDAR15 (right two columns) datasets.}
    \label{fig:res}
\end{figure*}

The results are shown in Tab.~\ref{tab:ex_ic15} and Fig.~\ref{fig:res}. Our method achieves 88.2\% F-Measure on the text detection task and is better than most previous methods. Besides, on the end-to-end recognition task, the PGNet-A with strong and weak lexicon achieves state-of-the-art accuracy. Particularly, our method exceeds the previous best one-stage text spotter CharNet R-50 by 3.2\%, 3.8\%, and 1.2\%  when evaluated with strong, weak, and general lexicon, respectively.

\begin{table*}
    \centering
    \begin{tabular}{l|c|c|c|c|c|c}
        \hline
        \multirow{2}{*}{Method} &\multirow{2}{*}{Backbone}  & \multicolumn{3}{c|}{Detection} & \multicolumn{1}{c|}{Recognition} & \multirow{2}{*}{FPS} \\ \cline{3-6}
        & & Recall  & Precious  & F-score  & None                         &                      \\ 
        \hline
        FOTS~\cite{liu2018fots}                 &ResNet-50          &38.0 &52.3 &44.0 &32.2    &-\\
        Textboxes~\cite{liao2017textboxes}      &ResNet-50-FPN      &45.5 &62.1 &52.5 &36.3    &1.4\\
        Mask TextSpotter~\cite{yao2018mask}     &ResNet-50-FPN      &55.0 &69.0 &61.3 &52.9    &4.8\\
        TextNet~\cite{sun2018textnet}           &ResNet-50-SAM      &59.5 &68.2 &63.5 &54.0   &2.7\\
        TextSnake~\cite{long2018textsnake}      &VGG16-FPN          &74.5 &82.7 &78.4 &-      &-\\
        TextDragon~\cite{feng2019textdragon}    &VGG16-FPN          &75.7 &85.6 &80.3 &48.8   &-\\
        CharNet H-57~\cite{xing2019convolutional}&Hourglass-57      &81.0 &88.6 &84.6 &63.6   &1.2\\
        Boundary~\cite{wang2020all}    &ResNet-50-FPN      &85.0 &\textbf{88.9} &\textbf{87.0} &\textbf{65.0}  &- \\
        ABCNet-F~\cite{liu2020abcnet}           &ResNet-50-FPN      &-    &-    &-    &61.9   &22.8\\
        \hline
        PGNet-A             &ResNet-50-FPN          & \textbf{86.8} &85.3 & 86.1 & 61.7  &38.2\\
        PGNet-A with GRM    &ResNet-50-FPN          & \textbf{86.8} &85.5 & 86.1 & 63.1  &35.5\\
        PGNet-E             &EfficientNet-B0-FPN    & 85.0 & 84.6 & 84.8 & 58.4 &\textbf{46.7}\\
        PGNet-E with GRM    &EfficientNet-B0-FPN    & 84.9 & 84.7 & 84.8 & 60.5 &40.5\\
        \hline
    \end{tabular}
    \caption{Evaluation on Total-Text for detecting text lines of arbitrary shapes.}
    \label{tab:tt}
\end{table*}

\begin{table*}
    \centering
	  \begin{tabular}{l|c|c|c|l|c|c|c}
	  \hline
		\multirow{2}{*}{Method} & \multicolumn{3}{c|}{Detection} & \multirow{2}{*}{Method} & \multicolumn{3}{c}{E2E Recognition} \\
		\cline{2-4} \cline{6-8} & R & P & F & & S & W & G  \\
 		\hline
        EAST~\cite{zhou2017east}                & 78.3 &83.3 &80.7 & OpenCV 3.0~\cite{karatzas2015icdar}   &13.8 &12.0 &8.0 \\
        LOMO~\cite{Zhang2019CVPR}               & 81.0 &91.6 &86.0 & Deep TextSpotter~\cite{busta2017deep}         &54.0 &51.0 &47.0 \\
        Mask TextSpotter~\cite{yao2018mask}     & 81.0 &91.6 &86.0 & Mask TextSpotter~\cite{yao2018mask}            &79.3 &73.0 &62.4 \\
        TextNet~\cite{sun2018textnet}           & 85.4 &89.4 &87.4 & TextNet~\cite{sun2018textnet}                  &78.7 &74.9 &60.5 \\
        FOTS R-50~\cite{liu2018fots}            & 85.2 &91.0 &88.0 & FOTS R-50~\cite{liu2018fots}                   &81.1 &75.9 &60.8 \\
        TextDragon~\cite{feng2019textdragon}    &83.8  &\textbf{92.5} &87.9 & TextDragon~\cite{feng2019textdragon}           &82.5 &78.3 &\textbf{65.2} \\
        CharNet R-50~\cite{xing2019convolutional} &\textbf{88.3} &91.2 &\textbf{89.7} & CharNet R-50~\cite{xing2019convolutional}    &80.1 &74.5 &62.2 \\
        Boundary~\cite{wang2020all}           & 87.5 &89.8 &88.6 & Boundary~\cite{wang2020all}                    &79.7 &75.2 &64.1 \\
        TextPerceptron~\cite{cheng21text}  & 82.5 &92.3 &87.1 & TextPerceptron~\cite{cheng21text}  &80.5 &76.6 &65.1 \\
        \hline
        PGNet-A             & 84.3 & 92.4 & 88.1 & PGNet-A & 82.9 & 77.7 & 62.3 \\
        PGNet-A with GRM    & 84.8 & 91.8 & 88.2 & PGNet-A with GRM &\textbf{83.3} &\textbf{78.3} & 63.5 \\
        PGNet-E             & 83.6 & 85.6 & 84.6 & PGNet-E & 80.2 & 74.9 & 57.4 \\
        PGNet-E with GRM    & 83.6 & 85.8 & 84.7 & PGNet-E with GRM & 80.5 & 75.3 & 58.7 \\
		\hline
	\end{tabular}
	\caption{Evaluation on ICDAR 2015 for detecting oriented text. ``P'', ``R'', ``F'' represent ``Precision'', ``Recall'', ``F-measure'' respectively. ``S'', ``W'', ``G'' represent recognition with ``Strong'', ``Weak'', ``Generic'' lexicon respectively.}
	\label{tab:ex_ic15}
\end{table*} 

\subsection{Runtime} 
In this paper, we propose a simple yet powerful arbitrarily-shaped text spotter, and considering the limitation of computing resources in different application scenarios, we propose PGNet-A and PGNet-E with different stem networks. The runtime of our method can be roughly divided into three parts: network inference stage, post-processing stage, and graph refinement stage. The runtime of PGNet on Total-Text is evaluated with NVIDIA Tesla V100-SXM2-16GB, which is the same as ABCNet. The longer side of the test image is resized to 640, and the batch size is set to 1 on a single GPU. It takes 15.1 ms,  6.4 ms, and 3.2 ms in the three stages, respectively. It is worth noting that the post-processing stage, which accounted for a large proportion of time cost, is executed with Python code and can be further optimized. In Total-Text, PGNet-E runs at 46.7 FPS with  84.8\% and 58.4\% F-Measure respectively on the detection and end-to-end recognition without lexicon, surpassing the previous arbitrarily-shaped text spotters in efficiency significantly, as depicted in Tab.~\ref{tab:tt}.

\section{Conclusion and Future Work}
In this paper, we propose a real-time arbitrarily-shaped text spotter PGNet together with a novel graph refinement module to relieve the inefficient problem of previous methods. Optimizing TCC maps with proposed PG-CTC loss, our PGNet is free from character-level annotations and gets rid of NMS and RoI operations in the inference stage which is single-shot, efficient and straightforward. Moreover, a novel graph refinement module is proposed to improve the end-to-end performance further by reasoning the word-level semantic information via a spatial graph neural network. The contributions of our paper are confirmed by comprehensive experiments and qualitative visualization in public benchmarks. In the future, we are interested in promoting the deployment of our novel real-time arbitrarily-shaped text spotter on  smart edge devices.

\section*{Acknowledgments}
This work is done when Pengfei Wang is an intern at Baidu Inc., and is also supported in part by the National Natural Science Foundation of China (NSFC) under Grant Nos. 61572387, 61632019, 61871304, and the Foundation for Innovative Research Groups of the NSFC under Grant 61621005.

\section{Appendix}
\subsection{Graph Convolution Layer}
A graph convolution layer takes as input a node feature matrix $\bm{X}$ together with an adjacency matrix $\bm{A}$ and outputs a transformed node feature matrix $\bm{Y}$.
Formally, a graph convolution layer in our case has the following formulation,
\begin{equation}
    \bm{Y} = \sigma([\bm{X}\|\bm{G}\bm{X}]\bm{W}),
\end{equation}
where $\bm{X} \in \mathbb{R}^{N\times d_{in}}$, $\bm{Y} \in \mathbb{R}^{N\times d_{out}}$, $N$ is the number of nodes, and $d_{in}$, $d_{out}$ are the dimension of input/output node features.
$\bm{G} = \bm{g}(\bm{X},\bm{A})$ is a symmetric normalized Laplacian matrix of size $N \times N$, and $\bm{g(\cdot)}$ is a function of $\bm{X}$ and $\bm{A}$. 
Operator $\|$ represents matrix concatenation along the feature dimension.
$\bm{W}$ is the learnable weight matrix with size $2d_{in} \times d_{out}$, and $\sigma(\cdot)$ is the non-linear activation function. In our paper, mean aggregation is adopted as $\bm{g(\cdot)}$ for the aggregation operation, 
\begin{equation}
    \bm{G}=\bm{\Lambda^{-\frac{1}{2}}}\bm{A}\bm{\Lambda^{-\frac{1}{2}}},
\end{equation}
where $\bm{\Lambda}$ is a diagonal matrix with $\bm{\Lambda}_{ii}=\sum_j \bm{A}_{ij}$. The topological structure is implicit in the adjacency matrix, and $\bm{A}_{ij}$ indicates the affinity between the ${i}$-th node and ${j}$-th node. For a point sequence $\pi = \{ p_1, p_2, \dots, p_N \}$,  the adjacency matrix is defined as
\begin{equation}
    \bm{A_{ij}}=1 - D(p_i, p_j) / \max(A),
\end{equation}
where $D(p_i, p_j)$ is a L2 distance between $p_i$ and $p_j$, and the diagonal elements of the adjacency matrix is always 1, which means that each node is self-connected. 

\subsection{Visualization and Discussion}
\begin{figure}
    \centering
    \includegraphics[width=0.9\linewidth]{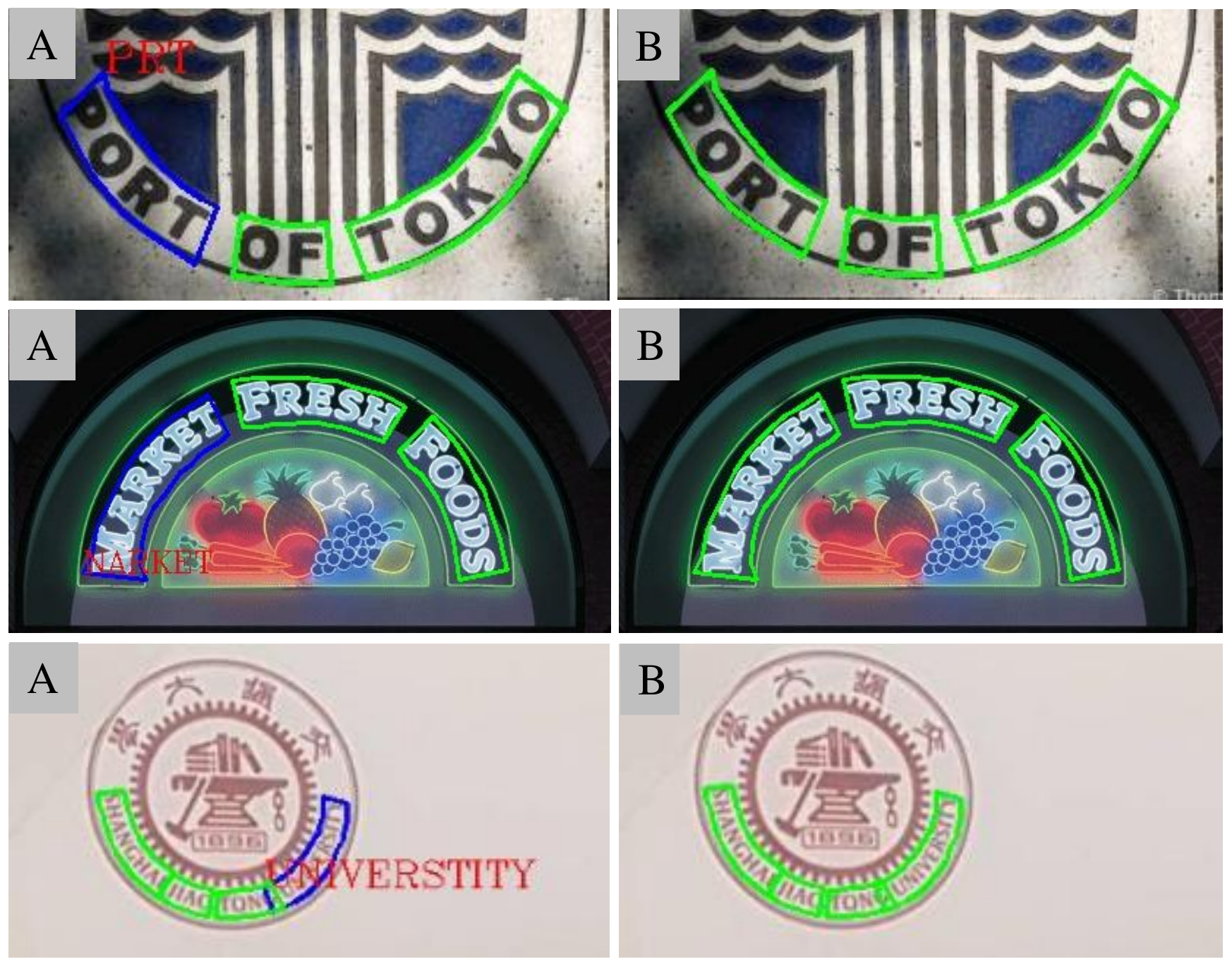}
    \caption{Qualitative results of our GRM to correct error recognition, where group A/B is the end-to-end results without/with GRM.}
    \label{fig:grm}
\end{figure}

\begin{figure}
    \centering
    \includegraphics[width=0.9\linewidth]{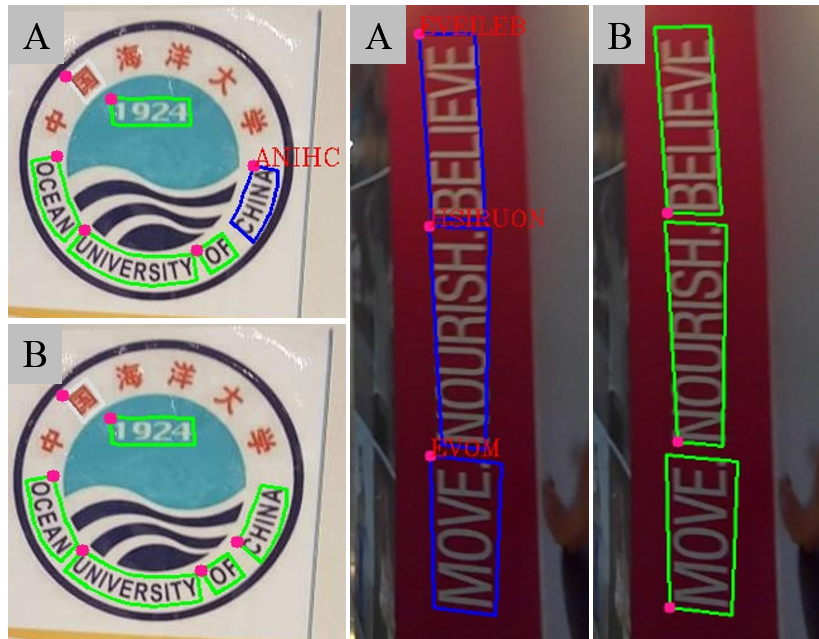}
    \caption{Qualitative results of our TDO to correctly recognize text in nontraditional reading directions, where group B/A is the end-to-end results with/without TDO.}
    \label{fig:tdo}
\end{figure}

We will do some visualization and case analysis to prove the effectiveness of PGNet and GRM qualitatively. In Fig.~\ref{fig:res},  Fig.~\ref{fig:grm} and Fig.~\ref{fig:tdo}, a green polygon means that detection and recognition both hit the ground truth. A blue polygon with red text indicates that the detection result hits, but the recognition result is incorrect. The right recognition results are not displayed for clear visualization. Besides, the upper left vertex of the text polygon in the right reading order is marked with a pink dot, if it exists. 

\textbf{Qualitative results of PGNet.} The visualization of spotting results in ICDAR2015 and Total-Text are shown in Fig.~\ref{fig:res}. As can be seen, the proposed arbitrarily-shaped text spotter PGNet can handle curved and multi-oriented text regions well. The superiority of TDO is proved in Fig.~\ref{fig:tdo}. The reading order information of group A is recovered by predefined rules, which is the same as \cite{feng2019textdragon}. Furthermore, group B is yielded by TDO map. As shown in Fig.~\ref{fig:res}, the word ``MOVE'' is misidentified as ``EVOM'' with predefined rules, while is recognized correctly with the TDO map. 

\textbf{Qualitative results of GRM.} The capability of GRM is depicted in Fig.~\ref{fig:grm}, and group A and group B is the end-to-end results without/with GRM. We show that the GRM can correct the recognition results in different scenarios. In the first case, the wrong recognition ``PRT'' is revised to the correct ``PORT'', proving the ability of GRM to make up for missing characters. In the second case, the wrong recognition ``NARKET'' is refined to the correct ``MARKET'', indicating the capability of correcting the wrong recognition of characters caused by similar appearance. It is shown in the third case that the ability of GRM to delete extra characters by correcting ``UNIVERSTITY'' to ``UNIVERSITY''. The cases mentioned above show that the GRM can add, modify and delete the error characters, which fully prove the potential of the GRM.

\bibliography{PGNet.bib}
\end{document}